# A Graph Prompt Fine-Tuning Method for WSN Spatio-Temporal Correlation Anomaly Detection

*Miao Ye , Jing Cui , Yuan Huang , Yong Wang , Qian He , Jiwen Zhang*

*Abstract*—Anomaly detection of multi-temporal modal data in Wireless Sensor Network (WSN) can provide an important guarantee for reliable network operation. Existing anomaly detection methods in multi-temporal modal data scenarios have the problems of insufficient extraction of spatio-temporal correlation features, high cost of anomaly sample category annotation, and imbalance of anomaly samples. In this paper, a graph neural network anomaly detection backbone network incorporating spatio-temporal correlation features and a multi-task self-supervised training strategy of "pre-training - graph prompting - fine-tuning" are designed for the characteristics of WSN graph structure data. First, the anomaly detection backbone network is designed by improving the Mamba model based on a multi-scale strategy and inter-modal fusion method, and combining it with a variational graph convolution module, which is capable of fully extracting spatio-temporal correlation features in the multi-node, multi-temporal modal scenarios of WSNs. Secondly, we design a three-subtask learning "pre-training" method with no-negative comparative learning, prediction, and reconstruction to learn generic features of WSN data samples from unlabeled data, and design a "graph prompting-fine-tuning" mechanism to guide the pre-trained self-supervised learning. The model is fine-tuned through the "graph prompting-fine-tuning" mechanism to guide the pre-trained self-supervised learning model to complete the parameter fine-tuning, thereby reducing the training cost and enhancing the detection generalization performance. The F1 metrics obtained from experiments on the public dataset and the actual collected dataset are up to 91.30% and 92.31%, respectively, which provides better detection performance and generalization ability than existing methods designed by the method.

*Index Terms*— Anomaly Detection, Graph Neural Networks, Pre-training, Prompt Learning, Wireless Sensor Networks

## I. INTRODUCTION

WIRELESS Sensor Networks (WSN) are multi-hop, self-organizing, distributed networks composed of a large number of sensor nodes connected via wireless means[1].They feature flexible network configuration and node deployment, enabling the collection of various environmental data such as temperature, humidity, light, and sound. WSNs are widely applied in fields such as national defense and security[2],environmental monitoring[3], industrial inspection[4],healthcare[5],intelligent transportation[6],and other fields.

With the widespread application of WSN in numerous fields, due to the open wireless communication characteristics of WSN, the transmission of WSN monitoring data is susceptible to human attacks and natural environmental influences, potentially causing the transmitted data to deviate from normal patterns[7], thereby affecting the integrity and authenticity of WSN monitoring data. Additionally, sensor network nodes themselves may experience failures that cause abnormal data, thereby affecting the reliability of WSNs. Therefore, efficiently detecting and locating abnormal data generated by faulty sensor network nodes is crucial for ensuring the reliability and stability of WSN operations.

In WSNs, the multiple physical quantities monitored by each sensor node can be represented as multivariate time series, also referred to as multimodal time series data in the literature[8]. Typically, WSN data anomalies refer to values or patterns that deviate significantly from normal data patterns. For single-time-series data, common anomaly types[9] include: point anomaly, contextual anomaly, and collective anomaly. Point anomaly[10] refers to a single data point that deviates significantly from the overall distribution of similar data. For example, in indoor temperature monitoring, if sensor readings generally range around 22°C , and one reading suddenly spikes to 45°C , this outlier may indicate equipment failure or a sudden heat source, constituting a typical point anomaly; Contextual anomaly[11] refers to data points that do not conform to expected patterns within a specific context or time frame but conform to normal patterns in other contexts. For example, in temperature time-series data from a temperate city during summer, if the temperature on a particular day is only 10°C , while the temperatures on the preceding and following days remain between 25°C and 35°C , the temperature on that day can be considered a contextual anomaly;Collective anomaly[12] refers to a group of data points that exhibit abnormal patterns simultaneously, such as a sudden increase in sensor readings deviating from the normal pattern within a specific area. Although individual data points may appear normal when viewed separately, they collectively form a collective anomaly. Additionally, for multiple WSN time-series data, temporal correlations exist between different modal time-series data collected by a single WSN node and between different WSN nodes. For example, temperature time-series data and humidity time-series data collected by the same WSN node typically exhibit a negative correlation, which represents temporal correlation between different time-series data[13] ; In a fire scenario, multiple temperature time-series data measured by different WSN nodes, and different temperature time-series data collected by multiple WSN nodes near the fire source exhibit a common upward trend and thus show a positive correlation. This correlation belongs to the spatial correlation between different time-series data[14]. Multiple WSN time-series data anomalies that violate these normal spatio-temporal correlation patterns can be referred to as correlation anomalies.

Based on the aforementioned anomaly types, researchers have proposed various anomaly detection methods by balancing anomaly detection accuracy and detection costs. Traditional anomaly detection methods include statistical-based methods and traditional machine learning methods, such as the autoregressive moving average model (ARMA)[15] , kernel density estimation (KDE)[16], Principal Component Analysis (PCA)[17] , and K-Means clustering[18]. These methods rely on prior assumptions about data distribution and feature extraction, and are limited by complex, dynamic structured data, making it difficult to fully extract the long-term dependency features and spatio-temporal correlation features of multi-node, multi-modal time-series data such as WSN. In contrast, deep learning methods possess the ability to automatically extract features, process large-scale data, and handle complex features. Commonly used methods include Convolutional Neural Networks (CNN)[19] , Recurrent Neural Networks (RNN)[20], and Long Short-Term Memory (LSTM)[21] are mainstream deep learning models that have been widely applied to anomaly detection tasks, enhancing the ability to model high-dimensional time-series data for anomaly detection. However, current approaches still face the following challenges when addressing anomaly detection tasks involving complex WSN time-series data:

First, deep learning-based WSN anomaly detection methods still have shortcomings in capturing complex WSN spatiotemporal dependencies. On the one hand, they are insufficient in capturing long-distance temporal dependencies, as reflected in the following: sequence modeling methods represented by RNN and its variants LSTM and Gated Recurrent Unit (GRU) perform well in modeling short-term dependencies, but when dealing with long-term dependencies spanning large time intervals, they often suffer from issues such as gradient vanishing or explosion, making it difficult to effectively transmit information over long distances. This prevents the model from fully retaining key historical features, thereby limiting its ability to understand and express complex temporal patterns. In WSN time-series data, abnormal behavior is often closely related to early states or trends. If such long-term dependencies are ignored, it may lead to a decrease in anomaly detection accuracy; On the other hand, most existing anomaly detection methods only consider the correlation between different modalities within the same [22] or the correlation between the same modality across different [23] when modeling the correlation in high-dimensional time-series data from WSNs. This fails to fully capture the complex correlation structures between multiple nodes and modalities, limiting their performance in addressing spatio-temporal correlation anomaly detection tasks.

Second, due to the complex spatio-temporal correlation features and long-term dependency characteristics of WSN time-series data, the data sample category annotation process is extremely challenging, and deep learning methods generally suffer from label deficiency issues. Supervised learning-based WSN anomaly detection methods heavily rely on sufficient and high-quality labels for training. When faced with large-scale unlabeled data, models often fail to adequately capture the true data distribution due to insufficient learning ability of normal patterns, thereby affecting model generalization performance and making it difficult to effectively identify unknown anomalies in real complex environments.

Finally, the performance of deep learning-based WSN anomaly node detection methods is highly dependent on the size of the sample set. In practical scenarios, anomaly events typically occur with extremely low frequency, inevitably leading to a shortage of anomaly samples. This situation constitutes a typical small-sample environment [22], where the model lacks sufficient anomaly samples during the training phase to effectively learn and capture the discriminative features of anomaly categories. Some literature has addressed the problem of few anomaly samples through resampling methods[24-25], but these methods have limitations: oversampling a small number of anomaly samples may lead to overfitting of the model to specific anomaly samples, while undersampling a large number of normal samples may result in the loss of important information, affecting the overall generalization performance of the model. Therefore, how to effectively improve the model's ability to recognize abnormal patterns in small sample environments with few abnormal samples is a key issue that must be considered when designing WSN abnormal node detection methods.

To address the aforementioned challenges, this paper proposes a multi-node, multi-modal sequential data anomaly detection model for WSN. It further introduces a learning method based on the "pre-training-graph prompting-fine-tuning" framework, effectively resolving issues such as missing training data labels and insufficient anomaly samples. The key innovations are as follows:

1) Existing WSN multi-temporal correlation anomaly detection methods are limited to single-node multi-modal or multi-node single-modal scenarios and are not applicable to multi-node multi-modal anomaly detection scenarios. Additionally, existing methods fail to adequately extract the spatio-temporal correlation features of WSN data. This paper designs an anomaly detection backbone network that integrates multi-temporal modal data from WSN and spatio-temporal correlation features between multiple nodes. By designing a multi-scale dilated convolution method to capture the local periodic patterns of WSN time-series data, combining Mamba to extract long-term time-series dependencies, and designing a cross-attention mechanism between different time-series modalities to extract their correlation features, while utilizing a variational graph convolution module to extract the spatial correlation features between nodes.

2) To address the challenge of high annotation costs associated with labeling actual WSN anomaly samples, which adversely affects anomaly detection performance, this paper proposes a self-supervised pre-training strategy based on three proxy sub-tasks: contrastive learning, prediction, and reconstruction. This approach enables the unsupervised learning of discriminative, sample-general feature representations.

3) To address the issue of sample imbalance caused by the scarcity of actual WSN anomaly samples, this paper proposes a "pre-training-graph prompting-fine-tuning" learning mechanism. This approach enhances anomaly detection capabilities under small-sample conditions while reducing training overhead. To the best of our knowledge, few existing studies have incorporated graph prompting mechanisms into WSN anomaly detection tasks.

## II. RELATED WORK

Since the data collected by each node in a WSN can be represented as multi-temporal data, this paper addresses the issues of spatio-temporal correlation anomalies, difficulty in labeling sample tags, and sample category imbalance caused by a small number of abnormal samples in the detection of anomalies in multi-temporal data in WSNs. In response, existing mainstream research has focused on two main directions. On one hand, studies have explored methods based on Multivariate Time Series Anomaly Detection (MTSAD), proposing detection frameworks that incorporate reconstruction, prediction, and self-supervised mechanisms to enhance the ability to capture temporal dynamics and inter-variable dependencies. On the other hand, with the development of pre-training and prompt learning (Prompt Learning) techniques, researchers have begun to explore their application in time series anomaly detection to enhance the model's knowledge transfer ability and generalization performance in low-resource scenarios. This section introduces the relevant work on WSN anomaly detection methods based on multi-time series data analysis and pre-training and prompt learning.

*A. Related Work on Multi-Variable Time-Series Anomaly Detection*

Compared with traditional single-variable detection methods, MTSAD not only needs to focus on the dynamic evolution of multi-variable time series but also needs to model the correlation features between variables, making the modeling and inference process more complex. Therefore, in recent years, based on the characteristics of WSN data that can be represented as multi-time series data, researchers have designed a variety of WSN anomaly detection methods based on MTSAD. Based on the minimization form of the loss function used in the training process[29], these methods can be categorized into three types: reconstruction-based, prediction-based, and self-supervised methods.

Reconstruction-based methods typically use an autoencoder structure to encode and decode input data, using reconstruction error as the anomaly score. Assuming that normal data can be reconstructed well during the training phase, abnormal data deviating from the normal pattern will have poorer reconstruction performance, resulting in higher anomaly scores. When the anomaly score exceeds a certain threshold, the data is deemed abnormal. DAEMON[30] is based on autoencoders and GAN structures. Autoencoders are used to reconstruct input time-series data, while GAN structures are used to constrain the intermediate outputs of autoencoders and the reconstruction outputs of autoencoders, respectively, to make the training process of autoencoder structures more robust and reduce overfitting. GReLeN[31] combines graph neural networks (GNN) and random graph relation learning within the VAE framework to capture dependencies between sensors and assign anomaly scores to each sensor based on reconstruction error. Deep Variational Graph Convolutional Recurrent Network (DVGCRN)[32] combines an embedding-guided probabilistic generative network with an adaptive variational graph convolutional recurrent network to model sensor dependencies and randomness in multivariate time series. MGCL[33] constructs a multi-order graph convolutional structure for joint modeling across variables and time windows, enabling robust perception of complex topological dependencies and temporal dynamics. It combines reconstruction tasks to extract global dependencies from multi-source sensor data.

Unlike reconstruction-based methods, which focus on measuring the error between input and reconstruction, prediction-based methods predict the future using historical windows and use prediction error as an anomaly score. Assuming that the model can accurately predict future values during the training phase, anomalies occur when the distribution shifts, causing prediction errors to increase significantly. If the prediction error exceeds a certain threshold, the value is deemed an anomaly. GTAD[34] combines graph attention networks and time convolutional networks to model both temporal dependencies and inter-variable correlations, using a single-dimensional prediction error as the loss function. Topo-GDN[35] introduces a topology analysis module and a multi-scale time convolutional module to the GDN framework, better preserving local node structures and global perception capabilities, and enhancing the modeling ability for complex spatio-temporal dependencies. PatchTST[36] reduces complexity through patching and utilizes channel independence to handle multi-modal time series. SageFormer[37] adds a global token to the beginning of each sequence based on Transformer to extract information from each variable sequence, and then uses graph learning to extract multi-variable relationships.

Although reconstruction and prediction methods have achieved certain results in practical applications, both rely on supervised learning mechanisms and loss function design that require explicit sample category labels. To further reduce dependence on labels and improve model generalization, self-supervised learning methods have gradually gained widespread attention in time-series anomaly detection in recent years. Their core advantage lies in their ability to fully mine the internal structure and potential patterns of data without relying on large amounts of labeled anomaly data. Self-supervised methods typically design auxiliary tasks such as predicting future time steps, reconstructing historical segments, and contrastive learning to learn transferable knowledge from large amounts of unlabeled data, thereby enhancing the model's generalization ability and extending it to downstream anomaly detection tasks. TranAD[26] proposes a bidirectional adversarial prediction framework based on Transformers, which constructs temporal constraints through forward and backward predictors and amplifies anomaly prediction residuals using a discriminator. CATCH[38] proposes a time-frequency joint optimization self-supervised learning framework that projects multivariate time series into the frequency domain and divides them into blocks to construct time comparison learning and variable-to-variable graph comparison learning objectives, thereby enhancing the model's ability to express periodic changes and variable-to-variable coupling features. VGATSL[39] combines variational graph attention and context contrast learning to simultaneously model sensor topology and temporal dependencies. TimeAutoAD[40] integrates topology-aware graph modeling with temporal information through self-supervised contrastive learning to construct robust feature representations.

Although the above WSN anomaly detection methods based on multi-temporal analysis have made some progress in reconstruction capability, prediction accuracy, and structural modeling, they still have limitations in handling correlation detection with complex spatiotemporal dependencies, labeling anomaly sample categories, and anomaly small sample problems. Specifically, due to the high dimensionality and complexity of multi-node, multi-modal, multi-temporal data, Most existing multi-temporal analysis methods focus on single-node multi-modal or multi-node single-modal WSN multi-temporal anomaly detection tasks, and it is difficult to fully extract the spatio-temporal correlation features between nodes and modalities in multi-node multi-modal scenarios. Given the adaptability and generalization advantages of self-supervised learning in low-label scenarios, this paper designs a multi-node multi-temporal anomaly detection method based on self-supervised mechanisms. By fully mining the spatio-temporal correlation features in WSN data, this method enhances the generalization capability of anomaly detection in multi-node multi-modal WSN scenarios.

*B. Pre-trained Models and Prompt Learning Related Work*

In recent years, "pre-training-fine-tuning" has become a core paradigm in fields such as natural language processing, computer vision, and graph representation learning. The basic idea is to perform pre-training on large-scale unlabeled data, combine auxiliary tasks to learn general representation capabilities, and then fine-tune the model with a small amount of data in downstream tasks. This effectively reduces the pressure of data annotation while improving model performance and generalization ability. For example, language models such as BERT[41] and GPT[42] can capture rich linguistic structures and semantic information; visual models such as ResNet and ViT[43] can recognize visual features such as colors, textures, and object edges. Currently, this paradigm is gradually attracting increasing attention in the field of time-series anomaly detection.

With the development of prompt learning, it has gradually been extended to non-linguistic domains, particularly graph representation learning. For graph-structured data, researchers have proposed graph prompt learning[44], which designs learnable graph prompt vectors for each graph or node to guide pre-trained graph models to focus on task-relevant sub-structures or adjacency patterns, and demonstrates its effectiveness in adapting to downstream tasks. Liu et al.[45] proposed a graph prompting framework that integrates pre-training and downstream tasks into a common task template, using learnable prompt vectors to bridge the gap between downstream tasks and pre-trained models, guiding downstream tasks to extract the most relevant knowledge from pre-trained models in a task-specific manner. Sun et al.[46] proposed GPPT, which converts independent nodes into token pairs and transforms downstream node classification into a link prediction problem through the combination of task tokens and structure tokens, and demonstrated its effectiveness through experiments. Fang et al.[47] proposed GPF, introducing a graph structure-aware prompt tuning mechanism. While freezing the main parameters of the pre-trained graph neural network, only a small number of graph prompt vectors are optimized, enabling flexible adaptation to tasks such as graph classification and node classification. Its efficiency and generalization ability under low-resource settings were validated on multiple benchmark datasets. Chen et al.[48] proposed GPL-GNN, which introduces prompt learning into graph neural networks, utilizing structure representations obtained from unsupervised pre-training as graph prompts to guide the model in extracting effective features in downstream tasks. This achieves lightweight fine-tuning without modifying the backbone network, enhancing adaptability to small-sample tasks.

Due to the graph structure characteristics of multi-node and multi-temporal WSN data, the spatial topological relationships between nodes are combined with multi-modal temporal signals within nodes, forming complex spatio-temporal dependency patterns. The traditional pre-training-fine-tuning paradigm often faces challenges such as difficulty in annotating downstream task samples and insufficient anomaly samples in anomaly detection tasks, making it difficult for pre-trained models to fully adapt to new anomaly detection environments. Graph prompt learning introduces learnable prompt vectors to guide the model to focus on node-local structures and temporal features relevant to anomaly detection while keeping the pre-trained model parameters frozen, effectively alleviating the distribution differences between the pre-training task and the anomaly detection task.

In summary, the "pre-training-fine-tuning" paradigm mainly relies on parameter adjustments to the pre-trained model to adapt to different downstream tasks, while the "pre-training-prompting" paradigm designs task-related prompt vectors to enable downstream tasks to flexibly adapt under fixed pre-trained model parameters, thereby significantly reducing fine-tuning costs and improving the model's generalization ability in low-resource environments. In multi-node, multi-modal temporal anomaly detection, the use of graph prompt learning not only fully utilizes the prior information of the spatial topology but also dynamically adjusts the weight of the attention to the temporal features of nodes through prompt vectors, thereby effectively improving the generality and accuracy of anomaly detection models.

III. PROBLEM DESCRIPTION AND BASIC PRINCIPLES

*A. Problem Definition*

Wireless sensor network (WSN) data is represented as $\mathbf{X} \in \mathbb{R}^{N \times M \times T}$, where $N$ denotes the number of sensor nodes, $M$ denotes the types of physical quantities (e.g., temperature, humidity, voltage, etc.) collected by each sensor node at any time $t$, and $T$ denotes the length of the collected time series. The data collected at time $t$ can be modeled as an attribute graph $\mathbf{G}_t = (\mathbf{A}, \mathbf{X}_t)$, where $\mathbf{A} \in \mathbb{R}^{N \times N}$ is the static adjacency matrix representing the adjacency relationships between any two nodes in the sensor network. When an edge exists between the $i$ th node and the $j$ th node, $\mathbf{A}(i,j)=1$; otherwise, $\mathbf{A}(i,j)=0$. $\mathbf{X}_t \in \mathbb{R}^{N \times M}$ is the attribute matrix of the nodes at time $t$, where the $i$ th row $x_i^t \in \mathbb{R}^M$ represents the multimodal time-series data of the node $i$ at time $t$. Therefore, the data collected over the entire time period $t = 1, 2, \ldots, T$ of the WSN can be represented as a graph-based time-series sequence:
$$\mathbf{G}_{1:T} = \{\mathbf{G}_1, \mathbf{G}_2, \ldots, \mathbf{G}_T\}, \text{ where } \mathbf{G}_t = (\mathbf{A}, \mathbf{X}_t)$$

Typically, an anomaly node detection model mapping function can be designed as $F$ and corresponding weight parameters $\theta$, with the input being a property graph subsequence. for a specific time period, resulting in:
$$\mathbf{Y} = F[\mathbf{G}_{t_1:t_2}; \theta] \tag{1}$$

$\mathbf{Y} \in \{0,1\}^{N \times M}$ represents the output of the detection task, where the $j$ th modal result $\mathbf{Y}_{i,j}$ of node $i$ can be represented as $y(i,j)$. If $y(i,j)=1$, it indicates that the WSN has an anomaly at node $i$ at the $j$ th modal, and if $y(i,j)=0$, it indicates that no anomaly occurred.

*B. Selective State Space Model*

To overcome the limitations of long-range dependency modeling and computational complexity, researchers have proposed a promising sequence modeling architecture called State Space Models (SSMs)[49], and developed efficient variants such as structured SSM[50] and Mamba[28][51]. The Mamba model combines selective SSM with efficient hardware-friendly algorithms and demonstrates outstanding performance across various modality tasks, such as language[28][51], image[52], medical[53], image data[54], recommendation systems[55], and time series analysis[56], demonstrating broad adaptability and robust modeling capabilities.

The theoretical foundation of the Mamba model originates from the continuous dynamic system description of state space models. Its mathematical modeling process follows the following core paradigm: given a continuous input signal $x(t) \in \mathbb{R}^n$, the system obtains an output signal $y(t) \in \mathbb{R}^m$ through nonlinear transformations of the hidden state $h(t) \in \mathbb{R}^d$. The computational process is as follows:
$$h'(t) = \mathbf{A}h(t) + \mathbf{B}x(t) \tag{2}$$
$$y(t) = \mathbf{C}h(t) \tag{3}$$

where $\mathbf{A} \in \mathbb{R}^{d \times d}$ is the state transition matrix, $\mathbf{B} \in \mathbb{R}^{d \times n}$ is the input projection matrix, and $\mathbf{C} \in \mathbb{R}^{m \times d}$ is the output projection matrix, all of which are learnable parameter matrices. To adapt to the requirements of discrete time series modeling, the continuous system can be converted into a discrete form (suitable for recursive forms in sequence modeling) using the zero-order holder (ZOH) discretization method:
$$\bar{\mathbf{A}} = \exp(\Delta \mathbf{A}) \tag{4}$$
$$\bar{\mathbf{B}} = (\Delta \mathbf{A})^{-1}(\exp(\Delta \mathbf{A}) - \mathbf{I}) \cdot \Delta \mathbf{B} \tag{5}$$
$$h_t' = \bar{\mathbf{A}}h_{t-1} + \bar{\mathbf{B}}x_t \tag{6}$$
$$y_t = \mathbf{C}h_t \tag{7}$$

where $\bar{\mathbf{A}}$, and $\bar{\mathbf{B}}$ are the discretization parameter matrices. This structure enables the model to process sequences with linear time complexity.

IV. GRAPH-PROMPTED FINE-TUNING-BASED WSN ANOMALY DETECTION METHOD

The anomaly detection method **GPamba-AD** (Graph-Prompted Mamba-based Anomaly Detection Model) designed in this paper, based on the "pre-training-prompt fine-tuning" paradigm, includes three aspects: data preprocessing strategy, feature extraction network, and model training strategy based on graph prompt fine-tuning. This section provides a detailed introduction to the implementation process of GPamba-AD as follows:

*A. Data Preprocessing*

Data collected from wireless sensor networks is typically recorded as observations from $N$ sensor nodes within a time interval $T$. Each node can simultaneously collect $M$ physical quantities, such as temperature, humidity, and light intensity, for detecting multiple modal indicators. Therefore, this data can be regarded as multi-node, multi-modal time-series data $\mathbf{X} \in \mathbb{R}^{N \times M \times T}$.

Since continuous time-series data is inherently a sequence of data with variable lengths, and most deep learning models require input in the form of fixed-length sequence samples, this paper adopts a sliding time window approach to divide the continuous time series into multiple time segments. This not only constructs more diverse samples during the training phase but also ensures temporal synchrony within the same window. Assuming the starting time step is $t_0$, the window length is $w$, and the sliding step size is $s$, the time segment within the starting window is represented as $[x_{t_0}, x_{t_0+1}, \ldots, x_{t_0+w-1}] \in \mathbb{R}^{N \times M \times w}$, and the next time segment is represented as $[x_{t_0+s}, x_{t_0+s+1}, \ldots, x_{t_0+s+w-1}] \in \mathbb{R}^{N \times M \times W}$.

For different physical quantities, differences in upper and lower limits may cause gradient explosion or disappearance, thereby affecting model training. It may also cause the model to misinterpret feature differences as anomalies, thereby affecting test accuracy. To eliminate the effects of different sensor units and numerical ranges, this paper uses the Z-Score method to standardize each time series of each node ($j \in [1, N]$) for each mode ($c \in [1, M]$) separately, as follows:
$$\mu_{j,c} = \frac{1}{W}\sum_{w=1}^{W} X_{w,j,c} \tag{8}$$
$$\sigma_{j,c} = \sqrt{\frac{1}{W}\sum_{w=1}^{W}(X_{w,j,c} - \mu_{j,c})^2} \tag{9}$$

$$\tilde{X}_{j,c} = \frac{X_{j,c} - \mu_{j,c}}{\sigma_{j,c} + \varepsilon} \quad (10)$$

where $\mu \in \mathbb{R}^{N \times M}$ represents the mean of the sequence data; $\sigma \in \mathbb{R}^{N \times M}$ represents the standard deviation of the sequence data; $\varepsilon$ represents a very small positive number to avoid division by zero; and $\tilde{X} \in \mathbb{R}^{N \times M \times W}$ represents the standardized output sequence.

This module is used to convert multi-node, multi-modal temporal data collected by wireless sensor networks into an input format suitable for model training and evaluation, ensuring that the model has stability and generalization capabilities when processing temporal data.

### B. Spatio-temporal Correlation Feature Extraction Network Design

Multi-node, multi-temporal data collected from wireless sensor networks has three dimensions: nodes, modalities (the physical quantities collected by each node, i.e., the feature dimensions of the temporal data), and time steps. To fully extract their feature information, this paper designs a spatio-temporal feature extraction network composed of a temporal feature extraction module and a spatial feature extraction module. The former is used to extract the temporal features of each node, while the latter mines the spatial correlation features between nodes by integrating spatial topology information. The specific architecture of the spatio-temporal feature extraction network is shown in Fig. 1.

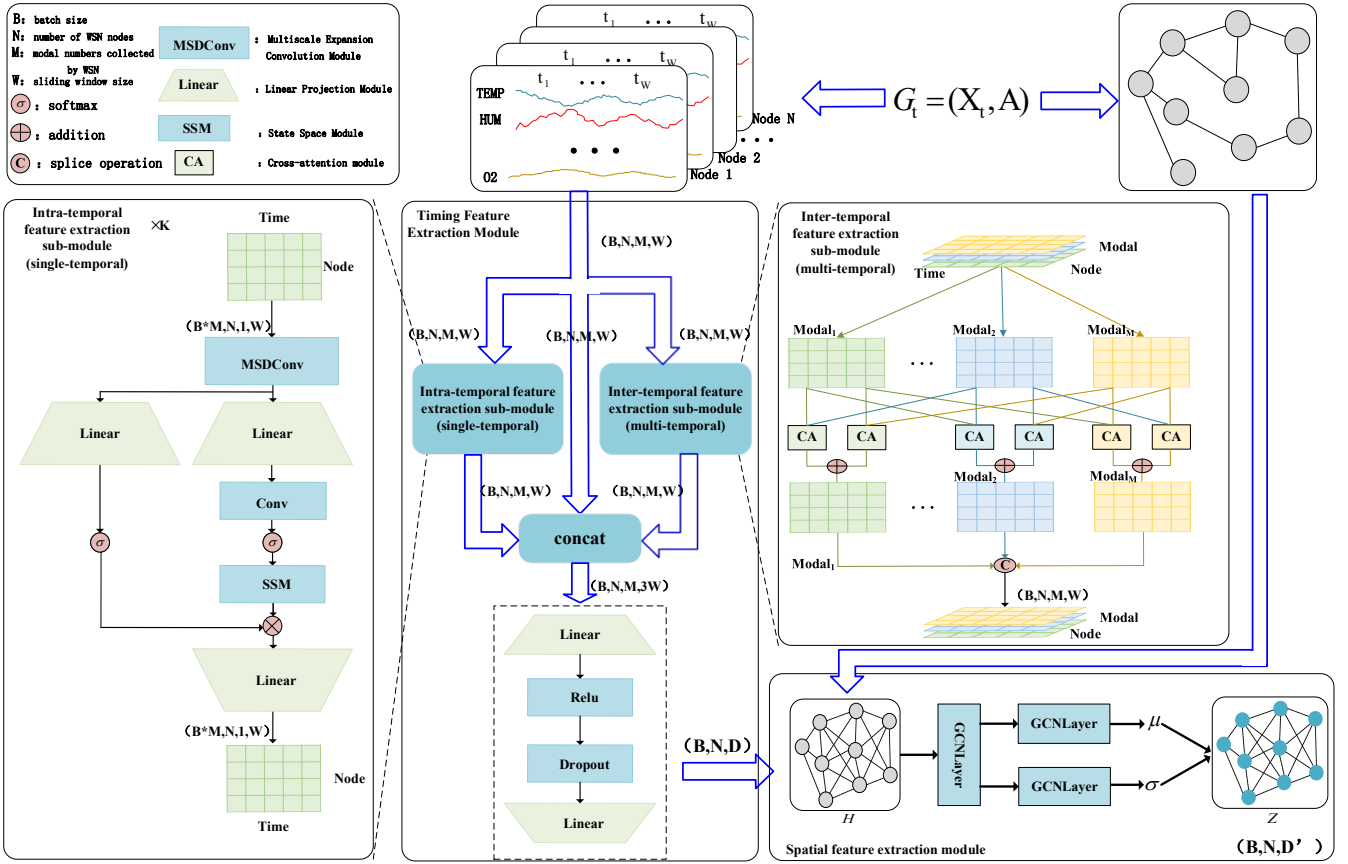

**Fig. 1.** Designed graph neural network anomaly detection backbone network with integrated spatio-temporal correlation feature

### 1. Temporal Feature Extraction Module

This module aims to extract key temporal feature information from the multi-node, multi-modal data $\mathbf{X} \in \mathbb{R}^{N \times W \times M}$ collected from WSN, including temporal inter-feature information (correlation features between different modalities) and temporal intra-feature information (dynamic representations of a single modality over time). By adopting a parallel structure design, two submodules are designed to extract features between different temporal modalities and temporal features within the same temporal modality, respectively.

To deeply extract feature information of a single temporal modality in the time dimension, this paper designs a temporal intra-feature extraction submodule based on the Mamba hierarchical structure. Each Mamba layer is composed of multi-scale dilated convolution (MSDConv) combined with the Mamba module to enhance the feature extraction capabilities for local fine-grained information and long-term dependencies.

Sequential data typically contains a mixture of short-term periodic fluctuations and long-term trend characteristics. While traditional sequential convolution methods can capture local dependency features, they are limited by their receptive field and struggle to extract long-range sequential features. To address this, the designed module employs multi-scale dilated convolution for local feature extraction, which expands the receptive field without sacrificing resolution. Additionally, the introduction of the Mamba module enhances global modeling capabilities, enabling the learning of long-range dependencies across time steps. The specific details are as follows: For single-modal time-series input $\mathbf{X}^{single} \in \mathbb{R}^{N \times W \times 1}$, first, multiple convolution kernels with different dilation rates are used to extract multi-scale features from the input as follows:

$$\mathbf{Z}^i = \text{Re}LU(\mathbf{W}_{d_i}\mathbf{X}^{single} + \mathbf{b}_i), i = 1, 2, ..., k \quad (11)$$

$$\mathbf{Z}^{ms} = Concat(\mathbf{H}^1, \mathbf{H}^2, ..., \mathbf{H}^k) \quad (12)$$

Among them, $\mathbf{W}_{d_i}$ denotes the weight matrix of the dilated convolutional layer with dilation rate $d_i$, $\mathbf{b}_i$ denotes the bias vector of the convolutional layer with dilation rate $d_i$, $k$ denotes the number of dilation rates, $Concat(\cdot)$ denotes concatenation along the feature dimension, $\mathbf{Z}^i \in \mathbb{R}^{N \times W \times 1}$ denotes the feature representation of the convolutional operation with dilation rate $d_i$, and $\mathbf{Z}^{ms} \in \mathbb{R}^{N \times W \times k}$ denotes the multi-scale feature representation.

The multi-scale feature representation $\mathbf{Z}^{ms}$ is then input into the Mamba module for global modeling. At time step $t$, the features are $z_t \in \mathbb{R}^{N \times k}$, dynamically generated through a linear projection layer:

$$\tilde{\mathbf{A}}_t, \tilde{\mathbf{B}}_t, \Delta_t = Linear_{SSM}(z_t) \quad (13)$$

where $\tilde{\mathbf{A}}_t \in \mathbb{R}^{N \times d}$ is the continuous state matrix parameter, $d$ is the latent state dimension, $\tilde{\mathbf{B}}_t \in \mathbb{R}^{N \times d}$ is the continuous input projection parameter, and $\Delta_t \in \mathbb{R}^{N \times 1}$ is the time step discretization parameter, used to control the state update frequency.

To convert the continuous system into a discrete-time system for temporal sampling, this paper applies zero-order hold (ZOH) to $\tilde{\mathbf{A}}_t$ and $\tilde{\mathbf{B}}_t$, as shown in formulas (16-19):

$$\overline{\mathbf{A}}_t = \exp(\Delta_t \tilde{\mathbf{A}}_t) \quad (14)$$

$$\overline{\mathbf{B}}_t = \Delta_t \tilde{\mathbf{B}}_t \quad (15)$$

$$h_t = \overline{\mathbf{A}}_t \cdot h_{t-1} + \overline{\mathbf{B}}_t \cdot z_t \quad (16)$$

$$y_t = \mathbf{C}_t \cdot h_t \quad (17)$$

where $\overline{\mathbf{A}}_t$ and $\overline{\mathbf{B}}_t$ are the corresponding discrete parameters of the continuous parameters $\tilde{\mathbf{A}}_t$ and $\tilde{\mathbf{B}}_t$, $h_t$ denotes the hidden state at the current time step, $h_{t-1}$ denotes the hidden state at the previous time step, $\mathbf{C}_t$ denotes the state output projection matrix, and $y_t$ denotes the output sequence at the current time step.

After two layers of temporal extraction, the final output sequence $\mathbf{Y} = \{y_1, y_2, ..., y_W\} \in \mathbb{R}^{N \times W \times M}$ is obtained.

To capture the random dependencies between different temporal modalities, this paper designs a multi-modal fusion temporal feature extraction submodule using a cross-attention mechanism. For data $\mathbf{X}_i \in \mathbb{R}^{N \times W}$ from modality $i$ and data $\mathbf{X}_j \in \mathbb{R}^{N \times W}$ from another different modality $j(j \neq i)$, the cross-attention mechanism is used to calculate their correlation and fuse them.

The sequence of modality $i$ is mapped to a query (Q), the sequence of another different modality $j(j \neq i)$ is mapped to a key (K) and a value (V), and the attention coefficients between modalities are calculated, as shown in formulas (20-23) below:

$$\mathbf{Q}_i = \mathbf{W}_Q \mathbf{X}_i \quad (18)$$

$$\mathbf{K}_j = \mathbf{W}_k \mathbf{X}_j \quad (19)$$

$$\mathbf{V}_j = \mathbf{W}_V \mathbf{X}_j \quad (20)$$

$$\alpha_{i,j} = softmax(\frac{\mathbf{Q}_i \mathbf{K}_j^T}{\sqrt{d_k}}) \quad (21)$$

Among them, $d_k$ represents the key-value dimension in the attention mechanism, $\mathbf{W}_Q$, $\mathbf{W}_K$, and $\mathbf{W}_V$ are learnable weight parameters, and the projection matrices $\mathbf{Q}_i \in \mathbb{R}^{N \times W \times d_k}$, $\mathbf{K}_j \in \mathbb{R}^{N \times W \times d_k}$, and $\mathbf{V}_j \in \mathbb{R}^{N \times W \times d_k}$ are obtained through matrix multiplication. $\alpha_{i,j} \in \mathbb{R}^{N \times W \times W}$ reflects the importance of modality $j$ when generating the output of the current modality $i$.

After obtaining the attention weights, the designed mechanism for fusing modal information at different time sequences can be represented as:

$$\mathbf{O}_i = \sum_{j=1}^{M} \alpha_{i,j} \mathbf{V}_j \quad (22)$$

$$\mathbf{O} = concat(\mathbf{O}_1, \mathbf{O}_2, ..., \mathbf{O}_M) \quad (23)$$

where $\mathbf{O}_i \in \mathbb{R}^{N \times W}$ represents the fusion result of information extracted from other modalities by modality $i$, and $\mathbf{O} \in \mathbb{R}^{N \times W \times M}$ represents the multi-modal fusion information.

Finally, to fuse the temporal dependency features of the single temporal modality with the characteristics of the multi-temporal modalities, this paper designs a parallel structure composed of an intra-temporal extraction submodule and an inter-temporal extraction submodule. The concatenates the outputs of these two modules, $\mathbf{Y}$ and $\mathbf{O}$, with the input data $\mathbf{X}$, and then performs feature compression and nonlinear mapping through a multi-layer perceptron (MLP), as shown in formulas (26-27):

$$\mathbf{F}_{concat} = concat(\mathbf{X}, \mathbf{Y}, \mathbf{O}) \in \mathbb{R}^{N \times W \times 3M} \quad (24)$$

$$\mathbf{F} = MLP(\mathbf{F}_{concat}) \in \mathbb{R}^{N \times W \times M} \quad (25)$$

This designed temporal feature extraction module not only achieves effective information fusion and dimension reduction but also preserves the detailed features in the original data. The module can model the long-term dependencies between different temporal sequences and the local dynamic changes and contextual features within a single temporal sequence, thereby generating more discriminative and robust temporal feature representations.

*2. Spatial feature Extraction Module*

Nodes in wireless sensor networks are typically distributed in a non-Euclidean spatial structure, making it challenging for traditional convolution operations to capture their adjacency relationships. To further model the spatial topology information between sensor nodes, this paper introduces a spatial feature extraction module after the temporal feature extraction module and designs a variational graph convolutional network (VGCN). By using reparameterization techniques to model the graph structure of multi-node temporal data, the module extracts potential spatial representations, and spatial dependencies between nodes are minded while preserving temporal features, thereby obtaining a complete spatio-temporal feature representation.

VGCN can explicitly model the probability distribution of node representations in the latent space, enabling the model to obtain deterministic feature embeddings while also modeling node uncertainty. This distributed representation helps improve the model's robustness to data perturbations, missing or abnormal samples, and is particularly suitable for wireless sensor network scenarios with high noise levels. Furthermore, VGCN maps graph data to a latent space, generating more compact and semantically rich node representations while preserving graph structural information. This process primarily utilizes GCN to perform a series of transformations on multi-node temporal features ($\mathbf{F}$) and the adjacency matrix ($\mathbf{A}$), with the specific calculation process shown below:

$$\mu = GCN_\mu(\mathbf{F}, \mathbf{A}) \qquad (26)$$

$$\sigma = GCN_\sigma(\mathbf{F}, \mathbf{A}) \qquad (27)$$

$$\mathbf{Z} = \mu + \sigma \odot \varepsilon \qquad (28)$$

where $GCN_\mu(\cdot)$ and $GCN_\mu(\cdot)$ represent the GCN layers with mean and variance outputs, respectively, and $\varepsilon \sim N(0,1)$ is the noise variable obtained from the standard normal distribution. The reparameterization formula(30) is used to obtain the latent embedding representation $\mathbf{Z}$.

This module aims to extract key temporal feature information from multi-node, multi-modal data collected from WSNs ($\mathbf{X} \in \mathbb{R}^{N \times W \times M}$), including inter-temporal features (correlation features between different modalities) and intra-temporal features (dynamic representations of a single modality over time). By adopting a parallel structure design, two sub-modules are designed to extract features between different temporal modalities and intra-temporal features within the same temporal modality.

*C. Design of a Graph-Prompted Fine-Tuning Training Strategy*

After designing the spatio-temporal correlation feature extraction network model for WSN multi-temporal data, to improve the performance of the WSN anomaly detection model, it is necessary to address the label missing problem and sample distribution imbalance caused by a small number of anomaly samples, which are commonly encountered in practical application scenarios. To address this, this paper proposes a graph-prompt fine-tuning training framework for the anomaly detection model designed earlier, as shown in Fig.2, which includes a graph pre-training module, a graph-prompt fine-tuning module, and an anomaly detection module, as described below.

*1. Design of the Graph Pre-training Module*

Given the characteristics of multi-node and multi-modal data in WSN, this paper designs a node spatio-temporal encoder based on the backbone network designed earlier, combined with a self-supervised learning strategy to more effectively extract time- and space-related features from WSN data. By modeling the WSN data as a dynamic graph sequence evolving over time ($\mathbf{G}_t = (\mathbf{X}_t, \mathbf{A})$), we design and construct a multi-task training framework with the following three sub-task optimization strategies to comprehensively improve the model's feature extraction capabilities.

Subtask 1: Negative-Free Graph Contrastive Learning (Negative-Free Graph Contrastive Learning) based on BYOL [59]. Common negative-free contrastive learning is mostly symmetric in structure, usually processing two different augmented views through two encoding branches and aligning the embedded representations in the same training process to effectively improve the model's discrimination ability in unsupervised scenarios. However, symmetric methods have some limitations: first, the two encoders share the gradient propagation path, which may cause the model to get stuck in a local optimum in the early stages of training, leading to instability in the learning objective; second, when the number of nodes is limited or the data augmentation noise is large, it may cause node data feature distribution shifts or alignment difficulties.

To mitigate these issues, this paper employs a non-symmetric zero-shot graph contrastive learning strategy based on BYOL. By designing encoder paths with independently updated parameters, a momentum encoder is used to provide a stable target, and a predictor is combined to achieve a non-symmetric alignment learning strategy. This enables the model to effectively improve its robustness and stability under graph structure perturbations and multi-view representations without introducing negative examples. This method is particularly suitable for wireless sensor network scenarios with limited node scale, strong structural locality, and high data augmentation variability, providing more stable spatio-temporal feature representations for WSN graph-structured data.

This paper completes the learning of this subtask by designing two neural networks: an online network and a target network.

Online Network: Includes the projection head $f_\theta$ and the prediction head $q_\phi$, with parameters $\theta$ and $\phi$; the input is the original image $\mathbf{G}_t$, and the output is its feature representation:

$$\mathbf{H} = f_\theta(\mathbf{G}_t) \qquad (29)$$

$$q = q_\phi(\mathbf{H}) \qquad (30)$$

where $q$ is the output feature of the online network, and $H$ is the spatio-temporal feature representation of the original image data.

Target Network: Contains only the projection head $f_\xi$. The parameters $\xi$ are copied from the online network via momentum update (EMA), as follows:

$$\xi = m \cdot \xi + (1-m) \cdot \theta, \ m \in [0,1] \quad (31)$$

The WSN dynamic graph sequence $\mathbf{G}_t = (\mathbf{X}_t, \mathbf{A})$ is processed by temporal masking[57] and edge perturbation[58] to obtain $\mathbf{G}'_t = (\mathbf{X}'_t, \mathbf{A}')$. The augmented graph $\mathbf{G}'_t$ is input into the target network, and its feature representation is output as:

$$z' = f_\xi(\mathbf{G}'_t) \quad (32)$$

The loss function ensures the network learns stable and consistent representations by maximizing the cosine similarity of positive sample pairs, using the following formula:

$$L_{cont} = 2 - 2 \cdot \frac{\langle q, z' \rangle}{\|q\| \cdot \|z'\|} \quad (33)$$

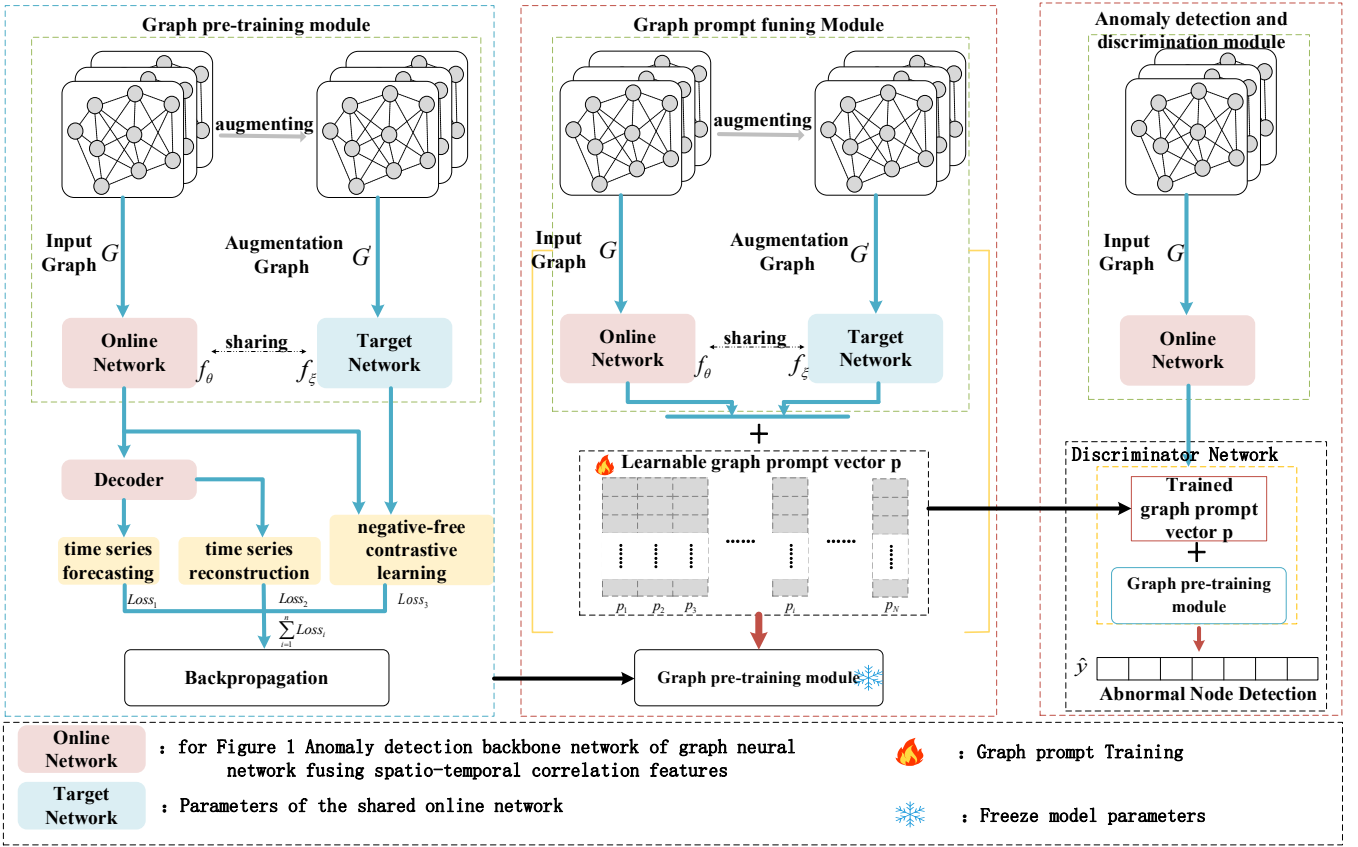

Fig.2. The proposed "pre-training-graph prompting-fine-tuning" training framework

**Subtask 2**: Multivariate temporal prediction subtask. This subtask utilizes historical spatiotemporal information to predict future node state information, enabling the encoder to effectively capture temporal evolution patterns and spatial structural relationships with predictive capabilities. We pass the encoded features obtained from the online network in Subtask 1, $\mathbf{H} = f_\theta(\mathbf{G}_t)$, through a multilayer perceptron (MLP) to predict the WSN data at the next time step, $\hat{\mathbf{Y}}_{t+1} = MLP(\mathbf{H})$. We use the mean squared error (MSE), a commonly used loss function in prediction models, to calculate the squared difference between the predicted values and the actual values, as shown in formula (36):

$$L_{pred} = \sum_{i=1}^{N} \sum_{j=1}^{M} [\hat{\mathbf{Y}}_{t+1}(i,j) - \mathbf{Y}_{t+1}(i,j)]^2 \quad (34)$$

where $i$ is the node index, and $j$ is the mode index.

**Subtask 3**: Multi-node temporal data reconstruction. Reconstruct the original data structure $\hat{\mathbf{X}}_{1:T} = MLP(\mathbf{H})$ from the encoded features $\mathbf{H} = f_\theta(\mathbf{G}_t)$ obtained in Subtask 1 using a multi-layer perceptron. The common mean square error (MSE) is used as the loss function in the reconstruction model, and the squared difference between the reconstructed values and the true values is calculated as shown in formula (37):

$$L_{recon} = \sum_{i=1}^{N} \sum_{j=1}^{M} \sum_{t=1}^{W} [\hat{X}_t(i,j) - X_t(i,j)]^2 \quad (35)$$

Among them, $i$ is the node index, $j$ is the modality index, and $t$ is the time step index.

By comparing the three figures above, designing and jointly optimizing the self-supervised sub-tasks of temporal prediction and temporal reconstruction, this paper constructs a unified multi-task training framework that can simultaneously capture WSN data structure-aware information, dynamic evolution patterns, and feature preservation capabilities, thereby fully extracting the spatio-temporal correlation features in WSN data.

Finally, the three loss functions from formulas (35–37) are jointly trained to form a new loss function as follows:

$$L = L_{cont} + L_{pred} + L_{recon} \quad (36)$$

Based on the aforementioned model design, the proposed spatio-temporal encoder is pre-trained using the backpropagation algorithm, and its parameters ($\Phi$) are learned and optimized to obtain the node representations ($h_1, h_2, ..., h_N$) in the wireless sensor network data. These feature representations and parameters obtained during the pre-training phase are then kept fixed in the subsequent prompt fine-tuning training phase.

**2. Graph Prompt Fine-tuning Module Design**

After obtaining the fixed model through the graph pre-training module, we adopt the method of embedding graph prompt vectors to enhance the adaptability of the pre-trained fixed model to downstream tasks. In this stage, we fix the parameters obtained from the pre-training model in Section 3.3.1 ($\Phi$), input the WSN data into its encoder to obtain node embedding vectors ($\mathbf{H} = f_{enc}(\mathbf{X}, \mathbf{A}) \in \mathbb{R}^{N \times d}$), and define a learnable graph prompt vector ($\mathbf{P} \in \mathbb{R}^{N \times d}$). For each node ($i$), we add the learnable prompt vector ($p_i \in \mathbb{R}^d$) to the embedding vector ($h_i \in \mathbb{R}^d$) of the corresponding node in Section 3.3.1, obtaining the prompt feature vector ($h_i^*$) of the generated vector ($h_i$) as shown below:

$$h_i^* = h_i + p_i \qquad (37)$$

By training the parameters $p$ in the pre-trained model, we can obtain the prompt features $\{p_1, p_2, ..., p_N\}$.

**3. Design of the anomaly detection classification module**

In the anomaly detection method based on multi-task learning designed in this paper, the prediction task best reflects the temporal dynamic changes of nodes. Therefore, only the future prediction task can be used as the basis for identifying abnormal nodes. After obtaining the fixed pre-trained model parameter matrix $\Phi$ and the trained prompt vectors $\{p_1, p_2, ..., p_N\}$, predict the temporal data of each node in the wireless sensor network at the next time step $\hat{Y}_{t+1} \in \mathbb{R}^{N \times M}$. The prediction error of each sensor node can be calculated as the anomaly score using the following formula:

$$score = \sum_{i=1}^{N} \sum_{j=1}^{M} [\hat{Y}_{t+1}(i,j) - Y_{t+1}(i,j)]^2 \qquad (38)$$

Compare the anomaly score $score$ with a threshold $\tau$. If the score exceeds the threshold, the node is deemed anomalous, and the label is set to 1; otherwise, it is set to 0:

$$label_{pred}(i,j) = \begin{cases} 1 & score(i,j) > \tau \\ 0 & score(i,j) < \tau \end{cases} \qquad (39)$$

Among them, $label_{pred}(i,j)$ represents the predicted label at sensor node $i$ in mode $j$, and $score(i,j)$ represents the anomaly score at sensor node $i$ in mode $j$.

## V. EXPERIMENTAL ANALYSIS

To validate the performance of the proposed graph-prompted fine-tuning WSN anomaly detection method, this paper conducts comprehensive testing on both public datasets and real-world datasets. This section first introduces the datasets and experimental environment configuration; then introduces the selected evaluation metrics; followed by ablation experiments to analyze the contribution of each module in the designed anomaly detection method to the overall detection performance; and finally compares the proposed method with existing mainstream methods to further validate its effectiveness; Finally, the model detection results are intuitively presented through visualization analysis.

### A. Experimental Dataset and Environment Configuration

The public dataset used in this paper is the IBRL (Intel Berkeley Research Lab Sensor Data) dataset deployed in the field by Intel Berkeley Lab, available at[60]. This dataset has been used in many literature[61], making it convenient for benchmark performance comparisons. The wireless sensor network consists of 54 sensor nodes. The laboratory collected four types of multi-modal information—temperature, humidity, light intensity, and voltage—over a period of more than one month in an indoor environment. The sampling time interval was 31 seconds, and the spatial distribution of the sensor nodes is shown in Fig.3. This public dataset was deployed in an indoor scenario. This paper also designed a WSN based on the LoRa communication protocol for outdoor scenarios to collect outdoor environmental data. It includes 14 sensor nodes for data collection and one data aggregation and transmission node. The relative spatial distribution of the sensor nodes and the physical diagram are shown in Fig.4. Similar to the IBRL dataset, each sensor node collects data including temperature, humidity, and node voltage every 30 seconds.

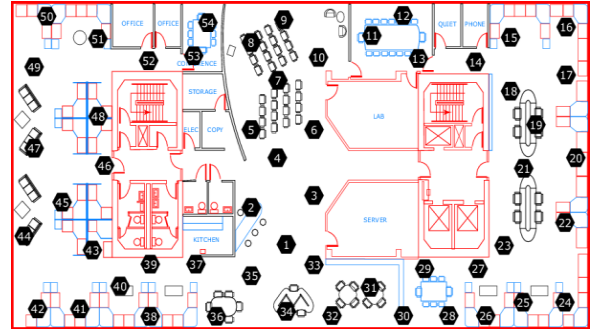

**Fig.3.** Sensor spatial distribution in the IBRL dataset

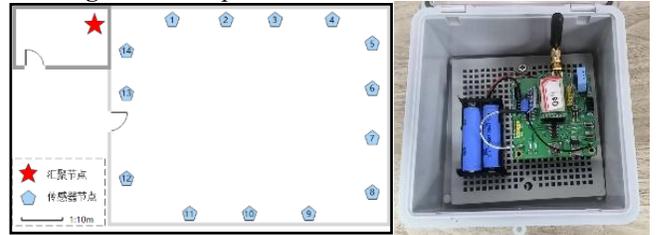

(a) Node distribution diagram    (b) Physical diagram of sensor nodes

**Fig.4.** Sensor node deployment of the WSN anomaly detection system

The specific hardware configuration of the experimental machine used for performance testing is as follows: Intel(R) Xeon(R) Gold 5218 CPU @ 2.30GHz, NVIDIA GeForce RTX 3090 GPU (24GB). The software environment includes: Ubuntu 18.04.2 LTS, Python 3.10, PyTorch 2.1.1, CUDA Version: 11.8. The source code was written by remotely connecting to the server via Pycharm 2022. The specific

4process of designing the model involving hyperparameter settings is as follows: For the batch size, we tried 8, 16, and 32. For the learning rate, we tried 0.001, 0.003, 0.005, 0.01, and 0.02. After multiple rounds of debugging, we found that the best results were achieved when the batch size was set to 16 and the learning rate was 0.005. Therefore, all the following experimental results are based on the above hyperparameter configuration (batch size = 16, learning rate = 0.005). Additionally, the sliding window size was set to 300, and the network model was optimized using the Adamw optimizer. The above hyperparameter settings were determined through multiple rounds of experimental tuning.

Similar to other literature studies, this paper selects precision (Pre), recall (Rec), and F1 score (F1) as evaluation metrics to assess the performance of the anomaly detection model.

*B. Ablation experiments*

To validate the performance and effectiveness of each component module in the proposed model and the impact of the self-supervised pre-training strategy on the overall model performance, a series of ablation experiments were designed. By removing or modifying various modules in the model, the contribution of each module to the overall performance can be comprehensively evaluated. The specific ablation experiment design is as follows:

**Scheme 1**: Use the original Mamba model without extracting fine-grained temporal feature information or fusing intermodal coherence features in the temporal domain.

**Scheme 2**: Use Mamba and the MSDConv module, without extracting multimodal correlation feature information between time series.

**Scheme 3**: Remove the VGCN module and do not extract node-to-node spatial correlation feature information.

**Scheme 4**: Remove the VGCN module and CA module, and do not perform any correlation feature information extraction.

**Solution 5**: Remove the self-supervised learning-based pre-training process and directly train for downstream tasks.

**Scheme 6**: Retain pre-training, but do not use prompt vectors during downstream task training; instead, directly train the anomaly detection task by freezing the pre-trained model parameters.

TABLE I
ABLATION EXPERIMENT SCHEMES AND RESULTS

| Method | MSDConv | CA | VGCN | Pretrain | GPL | Pre | Rec | F1 |
|---|---|---|---|---|---|---|---|---|
| 1 | × | × | √ | √ | √ | 89.30% | 69.87% | 78.40% |
| 2 | √ | × | √ | √ | √ | 88.64% | 82.98% | 85.71% |
| 3 | √ | √ | × | √ | √ | 81.25% | 82.98% | 82.11% |
| 4 | √ | × | × | √ | √ | 91.41% | 66.20% | 76.79% |
| 5 | √ | √ | √ | × | × | 89.69% | 88.83% | 89.26% |
| 6 | √ | √ | √ | √ | × | 93.18% | 87.23% | 90.11% |
| 7 | √ | √ | √ | √ | √ | **93.33%** | **89.36%** | **91.30%** |

The ablation experiment results of the anomaly detection method designed in this paper compared with the other four methods are shown in Table 1, where Pre denotes precision, Rec denotes recall, and F1 denotes the F1 score. Scheme 1 and Scheme 2 aim to explore the roles of different submodules in the temporal extraction module. Scheme 2 disables the modality fusion module, resulting in a 6.38% and 5.59% decrease in recall and F1 score compared to the baseline, indicating that the model's ability to detect anomalies in modality-related sequences significantly decreases when processing multimodal temporal data; Scheme 1 not only disables the mode fusion module but also the multi-scale convolution module, resulting in a 19.49% decrease in recall rate compared to the baseline. This indicates that the model fails to adequately extract local details from the temporal data, severely impacting detection performance.

Schemes 2, 3, and 4 focus on the role of the correlation extraction module. Scheme 3 disabled the VGCN module, resulting in a 9.19% decrease in the F1 score compared to the baseline. This was primarily due to the model's failure to detect anomalies in inter-node correlations, leading to overall detection performance degradation. Scheme 4 disabled both correlation extraction modules, further reducing performance. This indicates that, under the current dataset, detecting correlation anomalies is more challenging than detecting point anomalies, collective anomalies, and context anomalies, and the baseline model has relatively weaker recognition capabilities for such anomalies.

Schemes 5 and 6 analyze the impact on the model training framework. Scheme 5 removed the self-supervised pre-training process and directly performed the anomaly detection task, resulting in a 3.64% decrease in precision compared to the baseline. Scheme 6 retained pre-training but removed the prompt vectors. However, since the pre-training stage obtained general feature representations with good generalization from the WSN data, its precision improved by 3.49% compared to Scheme 5. Ultimately, the proposed complete solution achieves optimal performance across all metrics while retaining pre-training and introducing prompt vectors, further enhancing model performance.

*C. Comparison Experiments*

This section compares the **anomaly detection methods** used in this paper **with CNN-LSTM**[62], **MTAD-GAT**[27], **GAT-GRU**[63], and **GLSL**[63], and conducts the following comparison experiments.

**CNN-LSTM**[62] constructs a deep learning network utilizing convolutional modules, long short-term memory network modules, and fully connected layers.

**MTAD-GAT**[27] is an anomaly detection method that combines the strengths of predictive and reconstruction models.

process of designing the model involving hyperparameter settings is as follows: For the batch size, we tried 8, 16, and 32. For the learning rate, we tried 0.001, 0.003, 0.005, 0.01, and 0.02. After multiple rounds of debugging, we found that the best results were achieved when the batch size was set to 16 and the learning rate was 0.005. Therefore, all the following experimental results are based on the above hyperparameter configuration (batch size = 16, learning rate = 0.005). Additionally, the sliding window size was set to 300, and the network model was optimized using the Adamw optimizer. The above hyperparameter settings were determined through multiple rounds of experimental tuning.

Similar to other literature studies, this paper selects precision (Pre), recall (Rec), and F1 score (F1) as evaluation metrics to assess the performance of the anomaly detection model.

*B. Ablation experiments*

To validate the performance and effectiveness of each component module in the proposed model and the impact of the self-supervised pre-training strategy on the overall model performance, a series of ablation experiments were designed. By removing or modifying various modules in the model, the contribution of each module to the overall performance can be comprehensively evaluated. The specific ablation experiment design is as follows:

**Scheme 1**: Use the original Mamba model without extracting fine-grained temporal feature information or fusing intermodal coherence features in the temporal domain.

**Scheme 2**: Use Mamba and the MSDConv module, without extracting multimodal correlation feature information between time series.

**Scheme 3**: Remove the VGCN module and do not extract node-to-node spatial correlation feature information.

**Scheme 4**: Remove the VGCN module and CA module, and do not perform any correlation feature information extraction.

**Solution 5**: Remove the self-supervised learning-based pre-training process and directly train for downstream tasks.

**Scheme 6**: Retain pre-training, but do not use prompt vectors during downstream task training; instead, directly train the anomaly detection task by freezing the pre-trained model parameters.

TABLE I
ABLATION EXPERIMENT SCHEMES AND RESULTS

| Method | MSDConv | CA | VGCN | Pretrain | GPL | Pre | Rec | F1 |
|---|---|---|---|---|---|---|---|---|
| 1 | × | × | √ | √ | √ | 89.30% | 69.87% | 78.40% |
| 2 | √ | × | √ | √ | √ | 88.64% | 82.98% | 85.71% |
| 3 | √ | √ | × | √ | √ | 81.25% | 82.98% | 82.11% |
| 4 | √ | × | × | √ | √ | 91.41% | 66.20% | 76.79% |
| 5 | √ | √ | √ | × | × | 89.69% | 88.83% | 89.26% |
| 6 | √ | √ | √ | √ | × | 93.18% | 87.23% | 90.11% |
| 7 | √ | √ | √ | √ | √ | **93.33%** | **89.36%** | **91.30%** |

The ablation experiment results of the anomaly detection method designed in this paper compared with the other four methods are shown in Table 1, where Pre denotes precision, Rec denotes recall, and F1 denotes the F1 score. Scheme 1 and Scheme 2 aim to explore the roles of different submodules in the temporal extraction module. Scheme 2 disables the modality fusion module, resulting in a 6.38% and 5.59% decrease in recall and F1 score compared to the baseline, indicating that the model's ability to detect anomalies in modality-related sequences significantly decreases when processing multimodal temporal data; Scheme 1 not only disables the mode fusion module but also the multi-scale convolution module, resulting in a 19.49% decrease in recall rate compared to the baseline. This indicates that the model fails to adequately extract local details from the temporal data, severely impacting detection performance.

Schemes 2, 3, and 4 focus on the role of the correlation extraction module. Scheme 3 disabled the VGCN module, resulting in a 9.19% decrease in the F1 score compared to the baseline. This was primarily due to the model's failure to detect anomalies in inter-node correlations, leading to overall detection performance degradation. Scheme 4 disabled both correlation extraction modules, further reducing performance. This indicates that, under the current dataset, detecting correlation anomalies is more challenging than detecting point anomalies, collective anomalies, and context anomalies, and the baseline model has relatively weaker recognition capabilities for such anomalies.

Schemes 5 and 6 analyze the impact on the model training framework. Scheme 5 removed the self-supervised pre-training process and directly performed the anomaly detection task, resulting in a 3.64% decrease in precision compared to the baseline. Scheme 6 retained pre-training but removed the prompt vectors. However, since the pre-training stage obtained general feature representations with good generalization from the WSN data, its precision improved by 3.49% compared to Scheme 5. Ultimately, the proposed complete solution achieves optimal performance across all metrics while retaining pre-training and introducing prompt vectors, further enhancing model performance.

*C. Comparison Experiments*

This section compares the **anomaly detection methods** used in this paper **with CNN-LSTM**[62], **MTAD-GAT**[27], **GAT-GRU**[63], and **GLSL**[63], and conducts the following comparison experiments.

**CNN-LSTM**[62] constructs a deep learning network utilizing convolutional modules, long short-term memory network modules, and fully connected layers.

**MTAD-GAT**[27] is an anomaly detection method that combines the strengths of predictive and reconstruction models.

By modeling the temporal dependencies of data through its prediction module, this model can sensitively detect minute anomalous changes in time series, thereby providing more accurate references for subsequent reconstruction.

**GAT-GRU**[63] is a neural network architecture designed for multi-node, multi-modal sequential data. The model employs a reconstruction approach for anomaly detection: data is first mapped to a low-dimensional space, then restored to high-dimensional features via a symmetric structure, with the reconstruction error ultimately serving as the anomaly discrimination metric.

**GLSL**[63] is an enhancement of GAT-GRU that creates separate feature extraction branches for different modalities. By adjusting the computational order of the two GAT modules, it avoids structural redundancy in the model caused by an increase in the number of nodes or modalities.

TABLE II
COMPARISON EXPERIMENT RESULTS

| Method | Pre | Rec | F1 |
|---|---|---|---|
| CNN-LSTM | 79.5% | 70.0% | 74.5% |
| MTAD-GAT | 77.5% | 87.0% | 82.0% |
| GAT-GRU | 93.3% | 87.5% | 90.3% |
| GLSL | 94.5% | 87.0% | 90.6% |
| **ours** | **93.3%** | **89.4%** | **91.3%** |

The comparison results of the anomaly detection method designed in this paper with the other four methods are shown in TABLE II. It can be seen from the table that the experimental indicators of the CNN-LSTM method are lower than those of the other methods because this method uses CNN to perform convolution on multimodal time-series data, which can capture the features of single-modal time series but does not consider the complex correlation between different modal time series data, thereby limiting the performance of the model.

MTAD-GAT is primarily aimed at single-node anomaly detection tasks and does not consider the spatial dependencies between multiple nodes. In practical applications, this method typically requires training a separate model for each node to learn its temporal features, resulting in significantly increased computational costs and reduced efficiency in multi-node scenarios. Additionally, due to the lack of modeling capabilities for structural dependencies between nodes, MTAD-GAT has limitations in capturing cross-node anomaly propagation patterns, thereby affecting its detection performance in complex graph structures.

GAT-GRU and GLSL both model multi-modal time-series data features from three dimensions: time, modality, and space, and demonstrate good performance. GAT-GRU first independently extracts the corresponding time features and modality features within each node, then concatenates the feature vectors of all nodes for further space modeling across the entire graph. In contrast, GLSL divides multimodal data into multiple graph structures, extracts temporal and spatial features in parallel within each graph, and finally fuses representations from different modalities to model inter-modal correlations. In practical applications, since GAT-GRU requires an independent feature extraction module for each node, its computational overhead increases significantly with the number of nodes. Therefore, GLSL has better scalability and computational efficiency in large-scale graph data scenarios.

Compared with the existing representative method GLSL, the anomaly detection method proposed in this paper achieves better performance in terms of recall rate and F1 score. The experimental results show that the introduction of the graph prompt mechanism and multi-task optimization strategy can effectively enhance the model's comprehensive modeling ability for temporal patterns and spatial structure dependencies, thereby demonstrating better performance in anomaly node detection tasks.

TABLE III
COMPARISON RESULTS OF EXPERIMENTS UNDER DIFFERENT SCENARIOS

| Dataset | Pre | Rec | F1 |
|---|---|---|---|
| IBRL: Indoor Scene Public Dataset | 93.33% | 89.36% | 91.30% |
| Outdoor Scene Real-world Dataset | 93.82% | 90.85% | 92.31% |

To validate the generalization ability and stability of the proposed model, in addition to the IBRL public dataset collected in indoor scenes, this paper also collected actual data in outdoor scenes based on the LoRA communication method and verified it. The experimental results are shown in TABLE III. In indoor scenes, the anomaly detection method designed in this paper achieved 93.33% accuracy, 89.36% recall rate, and 91.30% F1 score, demonstrating good overall performance and indicating that the model can accurately identify anomalies. In the real-world dataset based on the LoRA communication method, the performance of the designed method was further improved, with an accuracy rate of 93.82%, a recall rate of 90.85%, and an F1 score of 92.31%. The experimental results from the two indoor and outdoor datasets jointly validated that the designed method demonstrates superior anomaly detection performance under various environmental conditions, fully demonstrating its excellent generalization ability and stability.

*D. Visualization Analysis*

This section conducts a case analysis of the model's detection performance using a portion of the test data. In Fig. 5, the red and green lines represent temperature and humidity data collected from node 28, containing 4,900 time points with added correlation anomalies. The black line indicates the true data labels, the blue line represents the predicted labels, and the orange line indicates misclassification cases. The visualization results of the correlation anomaly detection in this figure further validate the effectiveness of the individual components of the correlation detection model designed in this paper. Fig.6 shows a segment of temperature data containing 2,280 time points collected from node 27 in the IBRL dataset. The orange line represents the original normal data, the red markers indicate injected point anomaly data, the blue markers indicate injected context anomaly data, and the green markers indicate injected collective anomaly data. Test samples were constructed using a sliding window with a size of 300, and anomaly detection was performed based on the prediction results. Fig.7 shows the sample labels corresponding to different anomalies, and Fig.8 shows the labels of the anomaly detection results corresponding to different anomalies. It can be seen that the two are basically consistent, and the model misjudgments mainly occur when the sliding window just includes anomaly data at both ends and when the original normal data fluctuates greatly. The anomaly detection method designed in this paper successfully detected most of the anomaly data.

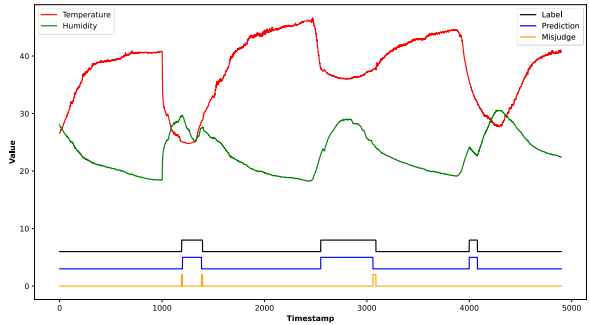

**Fig.5.** WSN data correlation anomaly detection

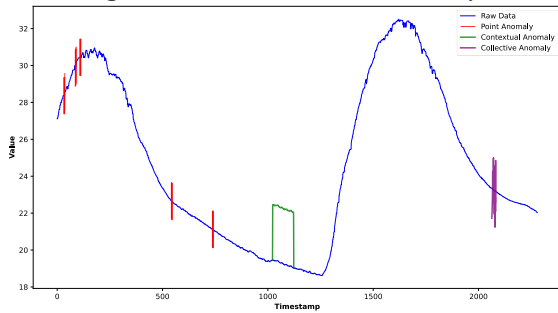

**Fig.6.** Test Sample Data

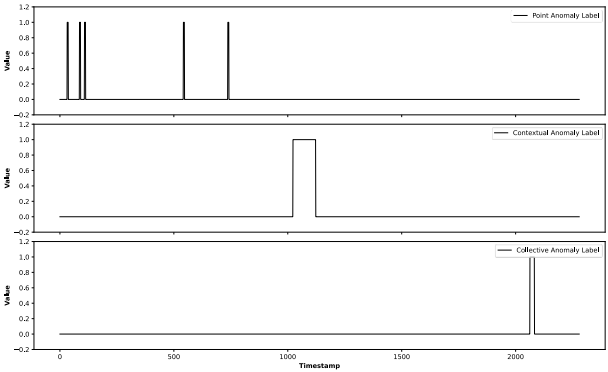

**Fig.7.** Test Sample Data Label

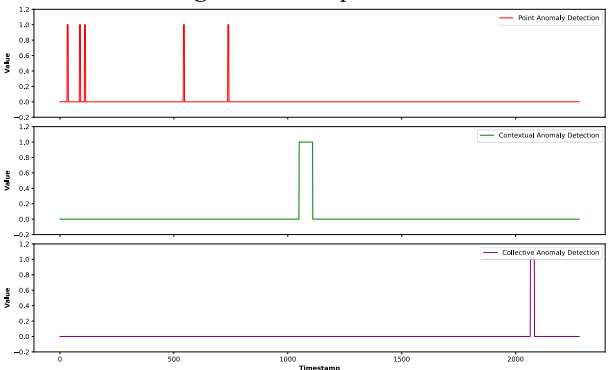

**Fig.8.** Test sample data anomaly detection results

## VI. CONCLUSION

This paper proposes a multi-node, multi-temporal anomaly detection framework based on pre-training and prompting fine-tuning, systematically integrating temporal modeling, graph structure modeling, and multi-task strategies. To address the issues of insufficient long-range dependencies and spatio-temporal relevance, this paper introduces a multi-scale dilated convolutional architecture combined with the Mamba module to enhance the modeling capability of multi-granularity temporal patterns within nodes. It also captures the relevance between different time sequences through a cross-attention mechanism and utilizes VGCN to obtain structural information from multiple nodes, thereby fully extracting high-quality spatio-temporal relevance features. To address the challenges of difficult sample labeling and a small number of anomaly samples, this paper designs a training strategy based on graph prompt fine-tuning. During the pre-training stage, the model simultaneously performs no-positive contrast learning, temporal reconstruction, and temporal prediction tasks, thereby enabling the model to learn latent representations with discriminative and generalizable capabilities. In downstream anomaly detection tasks, a small number of learnable graph prompt vectors are introduced. By freezing the pre-trained model and fine-tuning it on the prediction task, the model can adapt to specific scenarios with only a few parameter adjustments. Experimental results show that this method outperforms existing mainstream models in WSN anomaly detection tasks. In the next phase of research, we can consider how to further improve the feature extraction method, such as combining the time-frequency domain, time-sequence contrast learning, and graph contrast learning, to further improve the anomaly detection performance.

REFERENCES

[1] GULATI K, BODDU R S K, KAPILA D, et al. A review paper on wireless sensor network techniques in Internet of things (IoT) [J]. Materials Today: Proceedings, 2022, 51: 161-165.
[2] Behera T M, Mohapatra S K. A novel scheme for mitigation of energy hole problem in wireless sensor network for military application[J]. International Journal of Communication Systems,2021, 34(11):4886.
[3] Wang C, Shen X, Wang H, et al. Energy-efficient collection scheme based on compressive sensing in underwater wireless sensor networks for environment monitoring over fading channels[J]. Digital Signal Processing, 2022, 127: 103530.
[4] Dang T B, Le D T, Nguyen T D, et al. Monotone split and conquer for anomaly detection in IoT sensory data[J]. IEEE Internet of Things Journal, 2021, 8(20): 15468-15485.
[5] GAO C, YANG P, CHEN Y P, et al. An edge-cloud collaboration architecture for pattern anomaly detection of time series in wireless sensor networks[J]. Complex & Intelligent Systems, 2021, 7(5): 2453-2468.
[6] Ali T, Irfan M, Shaf A, et al. A Secure Communication in IoT Enabled Underwater and Wireless Sensor Network for Smart Cities [J]. Sensors, 2020, 20(15):4309.
[7] BOUBICHE D E, ATHMANI S, BOUBICHE S, et al. Cybersecurity issues in wireless sensor networks: current challenges and solutions[J]. Wireless Personal Communications, 2021, 117(1): 177-213.
[8] HUANG X H, ZHANG F B, FAN H Y, et al. Multimodal adversarial learning based unsupervised time series anomaly detection[J]. Journal of Computer Research and Development, 2021, 58(8): 1655-1667.
[9] IFZARNE S, TABBAA H, HAFIDI I, et al. Anomaly detection using machine learning techniques in wireless sensor networks[J]. Journal of Physics: Conference Series, 2021, 1743(1): 012021.
[10] S. Liu et al., "Time Series Anomaly Detection With Adversarial Reconstruction Networks," in IEEE Transactions on Knowledge and Data Engineering, vol. 35, no. 4, pp. 4293-4306, 1 April 2023, doi: 10.1109/TKDE.2021.3140058.
[11] Y. Jiao, K. Yang, D. Song, and D. Tao, "TimeAutoAD: Autonomous

Anomaly Detection With Self-Supervised Contrastive Loss for Multivariate Time Series," in IEEE Transactions on Network Science and Engineering, vol. 9, no. 3, pp. 1604-1619, May-June 2022, doi: 10.1109/TNSE.2022.3148276.

[12] Y. Li, X. Peng, J. Zhang, Z. Li, and M. Wen, "DCT-GAN: Dilated Convolutional Transformer-Based GAN for Time Series Anomaly Detection," in IEEE Transactions on Knowledge and Data Engineering, vol. 35, no. 4, pp. 3632-3644, April 1, 2023, doi: 10.1109/TKDE.2021.3130234.

[13] Ding Xiao'ou, Yu Shengjian, Wang Muxian, et al. Anomaly Detection of Industrial Time-Series Data Based on Correlation Analysis [J]. Journal of Software, 2020,31(03):726-747. DOI : 10 . 13328/ j.cnki.jos.005907

[14] Li G, Jung J J. Deep learning for anomaly detection in multivariate time series: Approaches, applications, and challenges[J]. Information Fusion, 2023, 91: 93-102.

[15] M. Celenk, T. Conley, J. Willis, and J. Graham, "Predictive Network Anomaly Detection and Visualization," in IEEE Transactions on Information Forensics and Security, vol. 5, no. 2, pp. 288-299, June 2010, doi: 10.1109/TIFS.2010.2041808.

[16] L. Zhang, J. Zhao, and W. Li, "Online and Unsupervised Anomaly Detection for Streaming Data Using an Array of Sliding Windows and PDDs," in IEEE Transactions on Cybernetics, vol. 51, no. 4, pp. 2284–2289, April 2021, doi: 10.1109/TCYB.2019.2935066.

[17] Pérez-Bueno F, García L, Maciá-Fernández G, et al. Leveraging a probabilistic PCA model to understand the multivariate statistical network monitoring framework for network security anomaly detection[J]. IEEE/ACM Transactions on Networking, 2022, 30(3): 1217-1229.

[18] E. Khaledian, S. Pandey, P. Kundu, and A. K. Srivastava, "Real-Time Synchrophasor Data Anomaly Detection and Classification Using Isolation Forest, KMeans, and LoOP," in IEEE Transactions on Smart Grid, vol. 12, no. 3, pp. 2378-2388, May 2021, doi: 10.1109/TSG.2020.3046602.

[19] SARANGI B, MAHAPATRO A, TRIPATHY B. Outlier detection using convolutional neural network for wireless sensor network[J]. International Journal of Business Data Communications and Networking, 2021, 17(2): 1-16.

[20] LAZAR V, BUZURA S, IANCU B, et al. Anomaly detection in software defined wireless sensor networks using recurrent neural networks[C]// Proceedings of the 2021 IEEE 17th International Conference on Intelligent Computer Communication and Processing (ICCP). Piscataway: IEEE Press, 2021: 19-24.

[21] MATAR M, XIA T, HUGUENARD K, et al. Multi-head attention based Bi-LSTM for anomaly detection in multivariate time-series of WSN[C]//Proceedings of the 2023 IEEE 5th International Conference on Artificial Intelligence Circuits and Systems (AICAS). Piscataway: IEEE Press, 2023: 1-5.

[22] Zeng F, Chen M, Qian C, et al. Multivariate time series anomaly detection with adversarial transformer architecture in the Internet of Things[J]. Future Generation Computer Systems, 2023, 144: 244-255.

[23] Wang X, Pi D, Zhang X, et al. Variational transformer-based anomaly detection approach for multivariate time series[J]. Measurement, 2022, 191: 110791.

[24] Xu H, Wang Y, Jian S, et al. Calibrated one-class classification for unsupervised time series anomaly detection[J]. IEEE Transactions on Knowledge and Data Engineering, 2024.

[25] Xu J, Wu H, Wang J, et al. Anomaly transformer: Time series anomaly detection with association discrepancy[J]. arXiv preprint arXiv:2110.02642, 2021.

[26] Tuli S, Casale G, Jennings N R. Tranad: Deep transformer networks for anomaly detection in multivariate time series data[J]. arXiv preprint arXiv:2201.07284, 2022.

[27] Zhao H, Wang Y, Duan J, et al. Multivariate time-series anomaly detection via graph attention network[C]//2020 IEEE international conference on data mining (ICDM). IEEE, 2020: 841-850.

[28] Gu, A.; and Dao, T. 2023. Mamba: Linear-time sequence modeling with selective state spaces. arXiv preprint arXiv:2312.00752.

[29] Kieu Khanh Ho T, Karami A, Armanfard N. Graph-based Time-Series Anomaly Detection: A Survey and Outlook[J]. arXiv e-prints, 2023: arXiv: 2302.00058.

[30] Chen X, Deng L, Huang F, et al. Daemon: Unsupervised anomaly detection and interpretation for multivariate time series[C]//2021 IEEE 37th International Conference on Data Engineering (ICDE). IEEE, 2021: 2225-2230.

[31] Zhang W, Zhang C, Tsung F. GRELEN: Multivariate Time Series Anomaly Detection from the Perspective of Graph Relational Learning[C]//IJCAI. 2022: 2390-2397.

[32] Chen W, Tian L, Chen B, et al. Deep variational graph convolutional recurrent network for multivariate time series anomaly detection[C]//International conference on machine learning. PMLR, 2022: 3621-3633.

[33] Cui Y, Zang D, Zhang J, et al. MGCL: Multi-Order Graph Neural Network with Cross-Learning for Multivariate Time-Series Anomaly Detection[J]. IEEE Transactions on Instrumentation and Measurement, 2025.

[34] Guan S, Zhao B, Dong Z, et al. GTAD: Graph and temporal neural network for multivariate time series anomaly detection[J]. Entropy, 2022, 24(6): 759.

[35] Liu Z, Huang X, Zhang J, et al. Multivariate time-series anomaly detection based on enhancing graph attention networks with topological analysis[C]//Proceedings of the 33rd ACM International Conference on Information and Knowledge Management. 2024: 1555-1564.

[36] Nie Y, Nguyen N H, Sinthong P, et al. A time series is worth 64 words: Long-term forecasting with transformers[J]. arXiv preprint arXiv:2211.14730, 2022.

[37] Zhang Z, Wang X, Gu Y S. Series-aware graph-enhanced transformers for multivariate time series forecasting. arXiv 2023[J]. arXiv preprint arXiv:2307.01616.

[38] Y. Xu, H. Wu, Y. Xie, J. Chen, G. Song, C. Wang, and J. Zhou, "CATCH: A context-aware and time-aware self-supervised framework for multivariate time series anomaly detection," in Proc. 29th ACM SIGKDD Int. Conf. Knowl. Discov. Data Min. (KDD), 2023, pp. 2234–2243, doi: 10.1145/3580305.3599419.

[39] Gao Y, Qi J, Ye H, et al. Variational Graph Attention Networks with Self-supervised Learning for Multivariate Time Series Anomaly Detection[J]. IEEE Transactions on Instrumentation and Measurement, 2024.

[40] Jiao Y, Yang K, Song D, et al. Timeautoad: Autonomous anomaly detection with self-supervised contrastive loss for multivariate time series[J]. IEEE Transactions on Network Science and Engineering, 2022, 9(3): 1604-1619.

[41] Devlin J, Chang M W, Lee K, et al. Bert: Pre-training of deep bidirectional transformers for language understanding[J]. arXiv preprint arXiv:1810.04805, 2018.

[42] Radford A, Narasimhan K, Salimans T, et al. Improving language understanding by generative pre-training[J]. 2018.

[43] Dosovitskiy A, Beyer L, Kolesnikov A, et al. An image is worth l6x l6 words: Transformers for image recognition at scale[J]. arXiv preprint arXiv:2010.11929, 2020.

[44] Sun Xiangguo, Zhang Jiawen, Wu Xixi, et al. Graph prompt learning: A comprehensive survey and beyond[J]. arXiv preprint, arXiv: 2311.16534,2023

[45] Liu Zemin, Yu Xingtong, Fang Yuan, et al. Graphprompt: Unifying pre-training and downstream tasks graph neural networks[C]//Proc of the ACM Web Conf 2023. New York: ACM, for 2023:417-428

[46] Sun Mingchen, Zhou Kaixiong, He Xin, et al. GPPT: Graph pre-training prompt tuning to generalize graph neural networks[C]//Proc of the 28th ACM SIGKDD Conf on Knowledge and Discovery and Data Mining. New York: ACM, 2022: 1717-1727

[47] Fang T, Zhang Y, Yang Y, et al. Universal prompt tuning for graph neural networks[J]. Advances in Neural Information Processing Systems, 2023, 36: 52464-52489.

[48] Chen Z, Wang Y, Ma F, et al. GPL-GNN: Graph prompt learning for graph neural networks[J]. Knowledge-Based Systems, 2024, 286: 111391.

[49] Gu, A.; Johnson, I.; Goel, K.; Saab, K.; Dao, T.; Rudra, A.; and Ré, C. 2021. Combining recurrent, convolutional, and continuous-time models with linear state space layers. Advances in neural information processing systems, 34: 572-585.

[50] Gu, A.; Goel, K.; and Ré, C. 2022. Efficiently Modeling Long Sequences with Structured State Spaces. In International Conference on Learning Representations.

[51] Dao, T.; and Gu, A. 2024. Transformers are SSMs: Generalized models and efficient algorithms through structured state space duality. arXiv preprint arXiv:2405.21060.

[52] Zhu, L.; Liao, B.; Zhang, Q.; Wang, X.; Liu, W.; and Wang, X. 2024. Vision mamba: Efficient visual representation learning with bidirectional state space model. arXiv preprint arXiv:2401.09417.

[53] Ma, J.; Li, F.; and Wang, B. 2024. U-mamba: Enhancing long-range dependency for biomedical image segmentation. arXiv preprint arXiv:2401.04722.

[54] Wang, C.; Tsepa, O.; Ma, J.; and Wang, B. 2024a. Graph-mamba: Towards long-range graph sequence modeling with selective state spaces. arXiv preprint arXiv:2402.00789.

[55] Yang, J.; Li, Y.; Zhao, J.; Wang, H.; Ma, M.; Ma, J.; Ren, Z.; Zhang, M.; Xin, X.; Chen, Z.; et al. 2024. Uncovering Selective State Space Model's Capabilities in Lifelong Sequential Recommendation. arXiv preprint arXiv:2403.16371.

[56] Liang A, Jiang X, Sun Y, et al. Bi-mamba4ts: Bidirectional mamba for time series forecasting[J]. arXiv e-prints, 2024: arXiv: 2404.15772.

[57] Luo D, Cheng W, Wang Y, et al. Time series contrastive learning with information-aware augmentations[C]//Proceedings of the AAAI Conference on Artificial Intelligence. 2023, 37(4): 4534-4542.

[58] You Y, Chen T, Sui Y, et al. Graph contrastive learning with augmentations[J]. Advances in neural information processing systems, 2020, 33: 5812-5823.

[59] Grill J B, Strub F, Altché F, et al. Bootstrap your own latent-a new approach to self-supervised learning[J]. Advances in neural information processing systems, 2020, 33: 21271-21284.

[60] Intel Berkeley Research Lab. IBRL Dataset[EB/OL]. (2004-06-02)[2025-05-26].https://db.csail.mit.edu/labdata/labdata.html.

[61] Akram J, Anaissi A, Akram A, et al. Adversarial Label-Flipping Attack and Defense for Anomaly Detection in Spatial Crowdsourcing UAV Services[J]. IEEE Transactions on Consumer Electronics, 2024.

[62] Dohare A K. A CNN and LSTM-based data prediction model for WSN[C]//2021 3rd International Conference on Advances in Computing, Communication Control and Networking (ICAC3N). IEEE, 2021: 1327-1330.

[63] Ye M, Zhang Q, Xue X, et al. A novel self-supervised learning-based anomalous node detection method based on an autoencoder for wireless

sensor networks[J]. IEEE Systems Journal, 2024, 18(1): 256-267.

.